%% file: main.tex
\documentclass[letterpaper, 10 pt, conference]{ieeeconf} 
\input{packages}

\IEEEoverridecommandlockouts 

\title{\LARGE \bf
Distilling 3D distinctive local descriptors for 6D pose estimation
}

\author{
Amir Hamza$^{*,\dagger}$,
Andrea Caraffa$^{*}$,
Davide Boscaini$^{*}$,
Fabio Poiesi$^{*}$\\
$^{*}$Fondazione Bruno Kessler, $^{\dagger}$University of Trento
\thanks{This study was funded by the European Union - NextGenerationEU, in the framework of the iNEST - Interconnected Nord-Est Innovation Ecosystem (Piano Nazionale di Ripresa e Resilienza (PNRR) – Missione 4, Componente 2, Investimento 1.5, D.D. 1058 23/06/2022, iNEST ECS00000043 – Spoke3, CUP E63C22001030007). The views and opinions expressed are solely those of the authors and do not necessarily reflect those of the European Union, nor can the European Union be held responsible for them. We acknowledge ISCRA for awarding this project access to the LEONARDO supercomputer, owned by the EuroHPC Joint Undertaking, hosted by CINECA (Italy)}
}

\begin{document}

\input{commands}

\maketitle
\thispagestyle{empty}
\pagestyle{empty}

\input{main/sections/0_abstract}
\input{main/sections/1_intro}
\input{main/sections/2_related}

\input{main/sections/3_prelim}
\input{main/sections/4_distill}
\input{main/sections/5_results}
\input{main/sections/6_conclusion}

{
    \small
    \bibliographystyle{IEEEtranBST/IEEEtran}
    \bibliography{main}
}

\end{document}

%% file: packages.tex
\usepackage{booktabs}
\usepackage{amsmath,amssymb,amsfonts}
\usepackage{multirow,multicol}
\usepackage{xspace}
\usepackage{cite}
\usepackage{url}
\usepackage{colortbl}
\usepackage{overpic}
\usepackage{microtype}
\let\labelindent\relax
\usepackage{enumitem}

\usepackage{xcolor}
\usepackage{booktabs}

%% file: commands.tex
\newcommand{\cmark}{\ding{51}}
\newcommand{\xmark}{\ding{55}}
\newcommand{\warning}[1]{\textbf{\color{red!90}{#1}}}

\definecolor{forestgreen}{RGB}{0,174,88}
\definecolor{bluediagram}{RGB}{0, 102, 204}
\definecolor{graydiagram}{RGB}{102, 102, 102}
\definecolor{reddiagram}{RGB}{255, 0, 0}
\definecolor{tableazure}{RGB}{214, 234, 248}

\newcommand{\fabiocomment}[1]{\todo[color=purple!20, inline, author=Fabio]{#1}}
\newcommand{\davide}[1]{\todo[color=blue!20, inline, author=Davide]{#1}}
\newcommand{\andreacomment}[1]{\todo[color=green!20, inline, author=Andrea]{#1}}
\newcommand{\amircomment}[1]{\todo[color=yellow!20, inline, author=Amir]{#1}}

\newcommand{\acronym}{dGeDi\xspace}
\newcommand{\eg}{e.g.,\xspace}
\newcommand{\ie}{i.e.,\xspace}

%% file: main/sections/0_abstract.tex
\begin{abstract}
Three-dimensional local descriptors are crucial for encoding geometric surface properties, making them essential for various point cloud understanding tasks.
Among these descriptors, GeDi has demonstrated strong zero-shot 6D pose estimation capabilities but remains computationally impractical for real-world applications due to its expensive inference process.
\textit{Can we retain GeDi's effectiveness, while significantly improving its efficiency?}
In this paper, we explore this question by introducing a knowledge distillation framework that trains an efficient student model to regress local descriptors from a GeDi teacher.
Our key contributions include:
an efficient large-scale training procedure that ensures robustness to occlusions and partial observations while operating under compute and storage constraints, 
and a novel loss formulation that handles weak supervision from non-distinctive teacher descriptors.
We validate our approach on five BOP Benchmark datasets and demonstrate a significant reduction in inference time while maintaining competitive performance with existing methods, bringing zero-shot 6D pose estimation closer to real-time feasibility.
Project website: \url{https://tev-fbk.github.io/dGeDi}.
\end{abstract}


%% file: main/sections/1_intro.tex
\section{Introduction}\label{sec:intro}

Three-dimensional descriptors (or features) encode geometric surface properties through compact numerical representations~\cite{rostami2019survey}. 
They are essential for various downstream tasks in point cloud understanding, including object 6D pose estimation~\cite{caraffa2024freeze}, registration~\cite{poiesi2023gedi}, segmentation~\cite{saltori2022gipso, garosi2025cops}, and navigation~\cite{Zhou2022}.
These descriptors can be broadly categorized as either \emph{local} or \emph{global}, depending on the type of information they capture.
Local descriptors operate at the point or patch level, capturing fine-grained geometric details~\cite{ao2021spinnet, wang2022you, poiesi2023gedi, boscaini2023patchmixer}.
Global descriptors also provide point-level representations, but aggregate geometric information from the entire point cloud~\cite{deng2018ppf, yew2022regtr, yu2024riga}.
Global descriptors are generally more computationally efficient~\cite{choy2019fcgf} but often sacrifice generalization across different data domains~\cite{poiesi2023gedi}.

Local descriptors can achieve rotation invariance through local reference frames (LRF)~\cite{poiesi2023gedi} or point-pair features (PPF)~\cite{deng2018ppfnet}. 
LRF-based methods establish a local coordinate system for each point using surrounding geometric information~\cite{toldi2017}. 
PPF-based methods compute antisymmetric 4D features from pairs of 3D points and their surface normals~\cite{ppf2010}.
Among local descriptors, GeDi~\cite{poiesi2023gedi} has been successfully applied to 6D pose estimation\footnote{6D pose estimation aims to recover the 3D position and orientation of an object in a scene from visual sensory data, such as RGBD images.} through the FreeZe method~\cite{caraffa2024freeze} in a zero-shot setting, achieving performances that, for the first time in the literature, nearly match those of state-of-the-art fully-supervised methods.
However, GeDi's inference time is impractical for real-world applications due to its computationally expensive process: LRF calculation\cite{toldi2017} for each point followed by a PointNet++-based descriptor encoding step~\cite{qi2017pn2} (Fig.~\ref{fig:teaser}, top).

\input{main/figures/teaser}

In this work, we explore how to address this inefficiency by using knowledge distillation~\cite{Hinton2014} to train a student model (Fig.~\ref{fig:teaser}, bottom) that regresses GeDi local descriptors, \ie the teacher model.
However, this presents several challenges. 
Distillation in this context requires large-scale point cloud datasets~\cite{chang2015shapenet,  downs2022google, deitke2024objaverse}. 
The rotation and scale invariance properties of the teacher model have to be effectively transferred to the student model.
Low inlier ratios must be handled, as only a small subset of descriptors is typically reliable, resulting in a significant imbalance during distillation, where most points provide unreliable supervision.
Despite these challenges, our approach presents unique opportunities. 
A more efficient descriptor extractor could bring zero-shot 6D pose estimation closer to real-time performance, making it more viable for practical applications.
Moreover, by tailoring the distillation process specifically for 6D pose estimation, we can improve descriptor robustness to occlusions and partial observations.
To the best of our knowledge, this is the first attempt to apply distillation to 3D local descriptors.

To this end, we introduce a distillation approach to train an efficient neural network (student) based on PointTransformerV3 (PTV3)~\cite{wu2024ptv3} to generate GeDi descriptors (teacher).
We refer to our approach as \acronym, derived from \underline{d}istilled \underline{GeDi}.
A key challenge in distilling GeDi is balancing efficiency and practicality.
On the one hand, a naive online distillation approach, where the teacher is computed jointly with the student, is impractical due to GeDi's slow run-time, making training excessively slow.
On the other hand, a naive offline distillation, where teacher features are precomputed, would be highly storage-inefficient.
To overcome these limitations, we propose a novel \emph{learning via correspondences} paradigm.
Offline, we precompute features from a query object, which is the 3D model, while online, we transfer them to the target object, the visible instance of this model present in the scene, using ground-truth transformations from synthetic 6D pose estimation datasets.
This allows scalable training on large datasets while keeping compute and storage requirements manageable.
Another challenge is that, unlike regression, where supervision comes from precise ground-truth annotations, our distillation supervision can include non-distinctive descriptors (\eg on the BOP Benchmark~\cite{2020_bop_report}, only 3\% of points can be matched through nearest neighbor search in the teacher feature space).
Standard loss functions~\cite{huberloss, focalloss} tend to overfit these outliers, producing unreliable student features.
To mitigate this, we introduce a novel loss function that weights each point's contribution based on its registration error. 
Points with errors above a certain threshold contribute almost equally, preventing excessive penalization from outliers.
Since training data are synthetic, we incorporate data augmentation to enhance rotational robustness in the student model and reduce the synthetic-to-real domain gap.
We validate \acronym through direct comparison with GeDi~\cite{poiesi2023gedi} and further evaluate its effectiveness by integrating it into FreeZe~\cite{caraffa2024freeze}, a state-of-the-art zero-shot pose estimator.
Additionally, we conduct an extensive ablation study to analyze \acronym's key components.
Experimental results demonstrate that \acronym significantly reduces inference time while maintaining competitive performance with alternative methods.
In summary, our main contributions are:
\begin{itemize}
    \item We introduce a distillation approach to learn local 3D descriptors for zero-shot 6D pose estimation using an efficient neural network, achieving a 170-times speed-up without compromising accuracy.
    \item We propose a data-scaling technique that enables training on large-scale synthetic data while significantly reducing memory requirements.
    \item We design a novel loss function that weights the regression error to prevent overfitting on synthetic data, thereby improving generalization.
\end{itemize}


%% file: main/figures/teaser.tex
\begin{figure}[t!]
    \centering
    \begin{overpic}[trim=30 15 26 1, width=0.95\columnwidth]{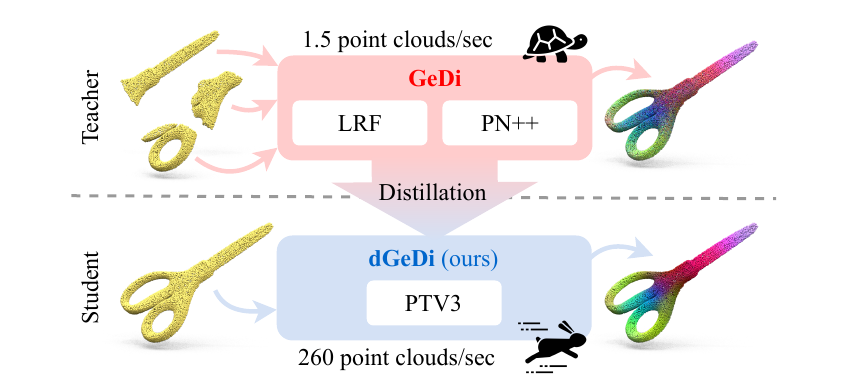}
    \end{overpic}
    
    \caption{
    We introduce \acronym, a 3D point cloud encoder trained by distilling GeDi~\cite{poiesi2023gedi} features (Teacher, Top) into a PointTransformerV3~\cite{wu2024ptv3} (PTV3) backbone (Student, Bottom).
    GeDi suffers from slow inference as it processes points sequentially (multiple input arrows), first extracting local reference frames (LRF) and then computing descriptors with PointNet++~\cite{qi2017pn2} (PN++).
    Instead, \acronym retains GeDi's generalization and distinctiveness while being over 170 times faster, making it ideal for real-time robotics applications.
    }
    \label{fig:teaser}
    \vspace{-1mm}
\end{figure}

%% file: main/sections/2_related.tex
\section{Related works}\label{sec:related}

\subsection{Zero-shot object 6D pose estimation}

Zero-shot pose estimation methods estimate the 6D pose of objects that are unseen during training.
These methods train on synthetic dataset or leverage the knowledge acquired by pre-trained vision and geometric foundational models.
The first category of methods like MegaPose~\cite{labbe2022megapose} and ZeroPose~\cite{chen2023zeropose} uses render and compare strategy.
At first a classification network selects the best-matching template from a set of rendered images that closely resemble the input image.
Then, a refiner network iteratively determines the pose by matching the selected template with additional viewpoints.
Since these methods requires a separate inference for each input image and template pair, the computational cost and inference time scale linearly with the number of templates.
Methods like GigaPose~\cite{nguyen2023gigapose} addresses the issue of inference time scaling. 
It renders templates to extract dense features using a Vision Transformer (ViT), and then finds the best template using fast nearest neighbour search in the feature space.
On top, two lightweight MLPs estimate the 2D scale and in plane rotation from a single 2D-2D correspondence and the two relative patch features extracted by another feature extractor.
SAM-6D~\cite{lin2023sam6d} leverages ViT to extract features that are back-projected into 3D to find 3D-3D correspondences between two point clouds. 
SAM-6D uses a matching network to enhance features before finding the actual correspondences.
FoundationPose~\cite{wen2023foundationpose} propose a generalizable and unified pipeline for pose tracking.
The coarse pose is estimated using a trained transformer based network.
This is followed by a pose refinement network that operates on input image crops and the renderings of the object model.
The refined poses are then ranked using a self attention based transformer network.

Other methods do not require task-specific training and leverage the knowledge learned by large scale vision and geometric foundation models.
ZS6D~\cite{ausserlechner2023zs6d} extracts features using pre-trained DINOv2~\cite{dinov2} from the inputs image and templates. 
Then, the best template is chosen using the cosine similarity and 2D-3D correspondences are retrieved using patch-wise similarity. 
The pose is estimated using PnP and RANSAC.
FoundPose~\cite{ornek2023foundpose} integrates DINOv2 patch features into bag-of-words representation for efficient template retrieval.
The pose hypothesis is generated using PnP and RANSAC and is refined using optimization-based photometric alignment algorithm.
FreeZe~\cite{caraffa2024freeze} fuses descriptors from pre-trained geometric and vision foundation models to extract discriminative features.
This is followed by RANSAC-based registration and a symmetry-aware pose refinement.

In this work, we use FreeZe’s geometric encoder as a teacher to train a more efficient student encoder via distillation.
This approach preserves the discriminative power of the original model while significantly improving efficiency.

\subsection{Feature distillation}
 
Knowledge distillation involves training a compact model using information from a more complex model or an ensemble of features \cite{hinton2015distilling}.
While this concept has been explored in various computer vision tasks, such as object detection~\cite{chen2017learning,chawla2021data,kang2021instance}, image recognition~\cite{huang2022feature,ma2022multi}, and segmentation~\cite{michieli2021knowledge,saltori2023pami}, its application to 6D pose estimation has been less explored.
Guo et al.~\cite{guo2023knowledge} introduce knowledge distillation for 6D pose estimation by supervising the student with the teacher's local predictions.
Both teacher and student networks process an RGB image to generate segmentation masks and local predictions, which are then used as correspondences for a PnP solver.
HRPose~\cite{guan2023hrpose} distills features from the teacher network using an efficient neural network architecture.
The student network is trained with supervision from both ground-truth 6D poses and geometric features via a feature similarity loss.
CleanPose~\cite{lin2025cleanpose} integrates causal learning with residual-based knowledge distillation to transfer rich category information from a pre-trained 3D encoder based on ULIP-2.
The frozen 3D encoder extracts a global point cloud descriptor, while the pose estimation network computes an average-pooled descriptor.

In this work, we distil knowledge from descriptors computed by a frozen teacher model, without relying on intermediate representations.
Instead, our student is trained to regress features, which are then used for pose computation via feature matching.
Like CleanPose, we use a frozen 3D encoder as the teacher, but instead of global CLIP-aligned representations, our approach uses rotation-invariant, highly distinctive point-level features from a local 3D encoder.

\subsection{Geometric encoders}

Geometric encoders map 3D points to high-dimensional feature representations that capture local geometric information. 
Existing methods operate at either \emph{point} or \emph{voxel} level.
\emph{Point-based} approaches process raw point clouds directly, either by encoding points independently~\cite{qi2017pn, qi2017pn2, qian2022pointnext} or within local neighborhoods to incorporate contextual information~\cite{GCNN, MoNet, SplineCNN, KPConv, wang2019dgcnn, boscaini2023patchmixer}.
\emph{Voxel-based} approaches first quantize point clouds into voxel grids at a given resolution and then efficiently process non-empty voxels using sparse 3D convolutions~\cite{choy20194d, choy2019fcgf} or space-filling curves~\cite{wu2024ptv3}. 
While voxelization improves scalability, it comes at the cost of fine geometric details.

In this work, we use a point-based encoder~\cite{poiesi2023gedi} as the teacher, which provides distinctive descriptor representations but incurs slower inference. 
Meanwhile, we employ a voxel-based encoder~\cite{wu2024ptv3} as the student, enabling faster runtime. 
The teacher relies on a custom module to estimate local reference frames (LRF in Fig.~\ref{fig:teaser}), ensuring rotation invariance by design. 
In contrast, the student follows a simple feed-forward architecture, requiring no custom modules, and achieves robustness through knowledge distillation.

%% file: main/sections/3_prelim.tex
\section{Preliminaries: FreeZe method} 
\label{sec:prelim}

Among existing zero-shot pose estimation methods in the literature~\cite{labbe2022megapose, lin2023sam6d, wen2023foundationpose}, we base our work on FreeZe~\cite{caraffa2024freeze}, which achieved state-of-the-art accuracy, winning the BOP Challenge 2024~\cite{bop2024challenge}.
FreeZe has three main stages: preprocessing, feature extraction, and feature matching.

\vspace{1mm}
\noindent\textit{Problem formulation.}
Given an RGBD image $\textbf{I} \in \mathbb{R}^{H \times W \times 4}$ capturing a scene and the 3D model $\mathcal{O}$ of an object within the scene, 6D pose estimation aims to determine the rigid transformation $\textbf{T}$ of $\mathcal{O}$ relative to the reference frame of the sensor that captured $\textbf{I}$.
Let $Q$ be the \emph{query} object, and $T$ be the \emph{target} object representing $\mathcal{O}$'s partial observation within $\textbf{I}$.
We define $\textbf{T}^{Q \to T} = ( \textbf{R}, \textbf{t} )$, where $\textbf{R} \in SO(3)$ is the rotation and $\textbf{t} \in \mathbb{R}^3$ is translation of $Q$.

\vspace{1mm}
\noindent\textit{Preprocessing.}
Let $\mathcal{P}^Q$ and $\mathcal{P}^T$ be the 3D point cloud of $Q$ and $T$, respectively.
$\mathcal{P}^Q$ is obtained by sampling points on $\mathcal{O}$'s surface. 
$\mathcal{P}^T$ is obtained by using an off-the-shelf zero-shot segmentor~\cite{lin2023sam6d} to extract the crop $\textbf{I}^T \subset \textbf{I}$ containing $T$, and then lifting $\textbf{I}^T$ in 3D using sensor's intrinsic parameters.

\vspace{1mm}
\noindent\textit{Feature extraction.}
FreeZe extracts features $\mathcal{F}_{}^Q$, $\mathcal{F}_{}^T$ from $\mathcal{P}^Q, \mathcal{P}^T$ using frozen vision~\cite{dinov2} and geometric~\cite{poiesi2023gedi} encoders, \ie $\mathcal{F} = ( \mathcal{F}_\text{geo} | \mathcal{F}_\text{vis} )$, where $\mathcal{F}_\text{geo}$ are the geometric features, $\mathcal{F}_\text{vis}$ are the visual features, and $|$ is the concatenation operator.

\vspace{1mm}
\noindent\textit{Feature matching.}
The transformation $\textbf{T}^{Q \to T}$ is estimated using RANSAC~\cite{RANSAC} and refined with ICP~\cite{chen1992object}. Point-level correspondences between $\mathcal{P}^Q$ and $\mathcal{P}^T$ are established via nearest neighbor search in the feature space.
Next, triplets of correspondences are sampled, and those leading to inconsistent alignments are discarded. For each valid triplet, the transformation is computed using Singular Value Decomposition (SVD), and the hypothesis with the highest inliers ratio is selected. Symmetries are solved using SAR~\cite{caraffa2024freeze}.


%% file: main/sections/4_distill.tex
\section{Object-oriented distillation}\label{sec:distillation}

\input{main/figures/diagram}

While FreeZe performs feature extraction with both geometric and vision encoders, we focus on the distillation of the former one\footnote{
Object-oriented distillation can also be applied to the vision encoder, but it is out of scope for this work.}, assuming $\mathcal{F} = \mathcal{F}_\text{geo}$.
We introduce a novel distillation procedure to transfer the distinctive properties of a slow teacher encoder to a faster student encoder (Fig.~\ref{fig:diagram}).
We define our proposed distillation approach as \emph{object-oriented} because training supervision is provided at the object level.
First, we extract teacher descriptors $\mathcal{F}^Q=\Phi_\Theta(\mathcal{P}^Q), \mathcal{F}^T=\Phi_\Theta(\mathcal{P}^T)$ using GeDi~\cite{poiesi2023gedi} encoder $\Phi_\Theta$.
Then, we learn student descriptors $\mathcal{G}^Q=\Psi_\Omega(\mathcal{P}^Q), \mathcal{G}^T=\Psi_\Omega(\mathcal{P}^T)$ with a PTV3~\cite{wu2024ptv3} encoder $\Psi_\Omega$.
During distillation, we optimize the parameters $\Omega$ 
so that $\mathcal{G}^Q \approx \mathcal{F}^Q$ and $\mathcal{G}^T \approx \mathcal{F}^T$, while $\Theta$ remain frozen.
Note that this differs from online knowledge distillation approaches~\cite{kdsurvey}, where both the teacher and student are neural networks and learn $\Theta, \Omega$ simultaneously.
%
%
In our case, the architectures of $\Phi_\Theta$ and $\Psi_\Omega$ are distinct (Fig.~\ref{fig:teaser}). Teacher features are precomputed, stored in memory, and loaded as needed during distillation.
To optimize this process, we propose storing teacher features only for query objects and introduce a module that leverages ground-truth 6D poses to transfer them to target objects (Sec.~\ref{sec:distill_corresp}).
Moreover, we propose a custom loss function that focuses learning on noise-free points, leading to improved performance (Sec.~\ref{sec:distill_loss}).

\subsection{Learning via correspondences}\label{sec:distill_corresp}
During training, given a data sample \((\mathcal{P}^Q, \mathcal{P}^T, \textbf{T}^{Q \to T})\), we leverage the ground-truth 6D pose \(\textbf{T}^{Q \to T}\) to align \(\mathcal{P}^Q\) with \(\mathcal{P}^T\) and perform a nearest neighbor search in the metric space to establish the correspondence $\Gamma \colon \mathcal{P}^T \to \mathcal{P}^Q$,
where each point \(\textbf{p} \in \mathcal{P}^T\) is mapped to  
$\textbf{q} = \operatorname*{argmin}_{\textbf{q} \in \mathcal{P}^Q} \lVert \textbf{T}^{Q \to T} \textbf{q} - \textbf{p} \rVert$, where \(\textbf{p}\) and \(\textbf{q}\) represent the 3D coordinates of corresponding points.
We use $\Gamma$ to transfer query teacher descriptors $\mathcal{F}^Q = \Phi_\Theta(\mathcal{P}^Q)$ to $\mathcal{P}^T$, obtaining $\tilde{\mathcal{F}}^T = \Gamma(\mathcal{F}^Q)$.
By replacing the explicit computation of $\mathcal{F}^T = \Phi_\Theta(\mathcal{P}^T)$
with $\tilde{\mathcal{F}}^T $
during distillation, we achieve two key benefits: (i) improved supervision for target objects and (ii) a scalable, object-oriented distillation process.

\vspace{1mm}
\noindent\emph{Improved target supervision.}
Matching features $\mathcal{F}^Q, \mathcal{F}^T$ is challenging because input data differ significantly: $\mathcal{P}^Q$ is computed from noise-free synthetic CAD models, while $\mathcal{P}^T$ is extracted from noisy real-world RGBD images.
This domain gap results in variations in point densities, partiality, and artifacts created by lifting the input image $\textbf{I}$ to generate $\mathcal{P}^T$.
For example, on the MegaPose-GSO~\cite{labbe2022megapose} dataset, we measured that, on average, only 21.4\% of the points can be matched via nearest neighbor search in the teacher feature space.
This means that 78.6\% of the features are considered outliers for feature matching.
In contrast, by transferring query teacher features to target point clouds with $\Gamma$, we simulate the ideal scenario where all target points are matched with the corresponding query points, promoting reliable supervision.

\vspace{1mm}
\noindent\emph{Improved scalability by reducing storage requirements.}
The MegaPose-GSO dataset consists of 850 query objects and 16M target objects, resulting in a 1:19k ratio.
This ratio can be further expanded by applying random transformations to the existing data.
As a result, storing target teacher features for a large-scale training dataset becomes prohibitively expensive in both computation and storage.
For example, the entire MegaPose~\cite{labbe2022megapose} dataset would require 14TB of storage.
In contrast, our approach significantly improves scalability by storing only the query teacher features, reducing the storage requirements for MegaPose by 411 times (\ie 34GB).


\subsection{Loss function design}\label{sec:distill_loss}

Although GeDi's descriptors have proven to be both generalizable and distinctive~\cite{poiesi2023gedi}, they still exhibits low inlier ratios, with an average RON~\cite{Zhou2022} of 3\% on the BOP Benchmark and 21.4\% on MegaPose-GSO.
This indicates that only a small fraction of points are highly distinctive, while the majority have ambiguous features.
Consequently, a loss function that assigns equal weight to all points will be dominated by outliers, leading to overfitting on ambiguous features.
To solve this issue, we propose a novel loss function:
\begin{equation}\label{eq:v2}
    \mathcal{L}(\mathcal{F}, \mathcal{G}) = 
    \begin{cases}
        \alpha \lVert \mathcal{F} - \mathcal{G} \rVert^2, & \text{if}\; \lVert \mathcal{F} - \mathcal{G} \rVert < \delta \\
        \beta \lVert \mathcal{F} - \mathcal{G} \rVert + \gamma, & \text{otherwise}.
    \end{cases}
\end{equation}
Eq.~\ref{eq:v2} is inspired by Huber loss~\cite{huberloss}, with a quadratic behavior for points with low regression error (below a threshold $\delta$) and a linear behavior otherwise.
In our experiments, we set $\alpha=8$, $\beta=1/5$, $\gamma = 8\delta^2 - \delta/5$, and $\delta=0.3$.
With these hyper-parameters, the quadratic component is relatively steep, while the linear component is nearly flat.
This design achieves two key objectives:
(i) for low regression errors, the significant variation in loss values encourages accurate regression;
(ii) for high regression errors, the minimal variation in loss values prevents excessive focus on outliers, assigning significantly smaller loss value compared to traditional loss designs such as MSE and Huber losses.
To validate this assumption, in Sec.~\ref{sec:quant_results} we present an experimental comparison with an alternative loss formulation that prioritizes learning from points with high regression error:
\begin{equation}\label{eq:v1}
    \tilde{\mathcal{L}}(\mathcal{F}, \mathcal{G}) = \lVert \mathcal{F} - \mathcal{G} \rVert ^\epsilon \mathcal{L}(\mathcal{F}, \mathcal{G}),
\end{equation}
where $\mathcal{L}$ is from Eq.~\ref{eq:v2} and $\epsilon$ is a hyper-parameter inspired by the Focal loss~\cite{focalloss}.
In our experiments, we set $\epsilon=15$ to guarantee that the points with large regression error receive significantly higher weights than the points with small error.

%% file: main/figures/diagram.tex
\begin{figure}[t!]
    \centering
    \hspace{-0.5cm} 
    \begin{overpic}[trim=14 0 0 0, clip, width=0.95\columnwidth]{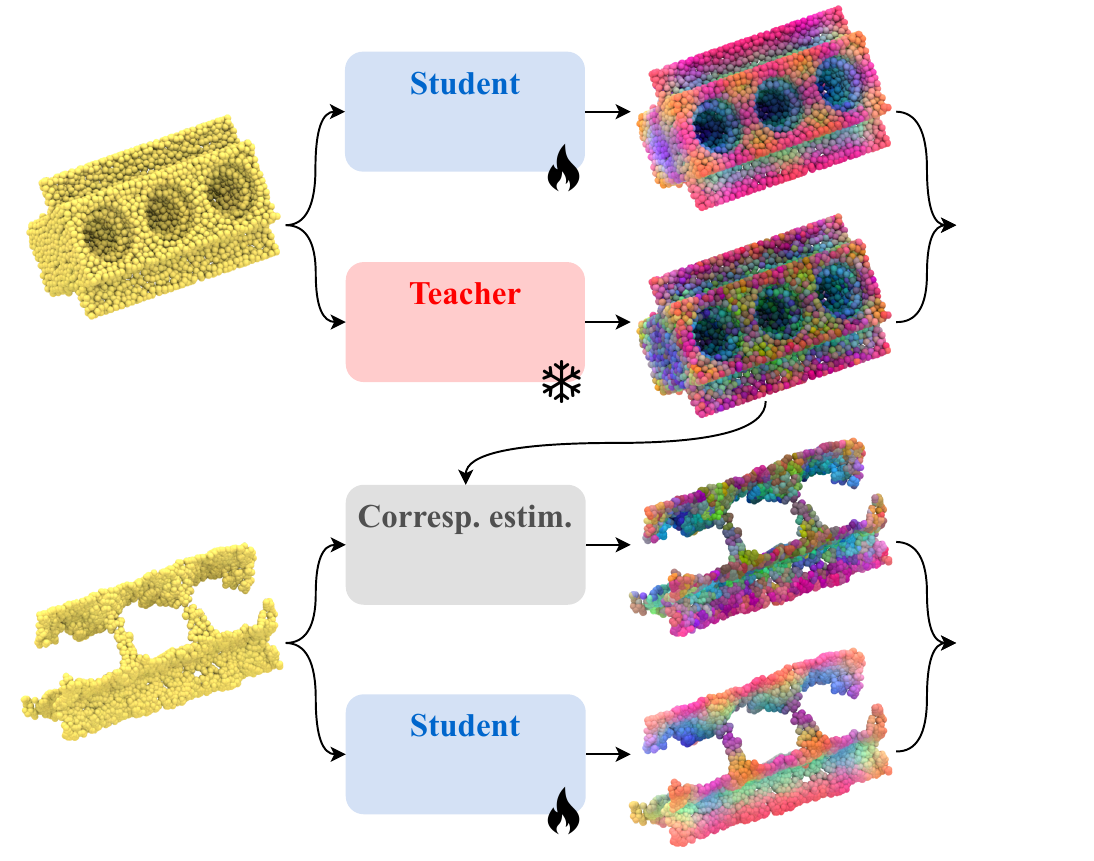}
    \put(18, 13){\footnotesize $\mathcal{P}^T$}
    \put(38.5, 7){\color{bluediagram}{$\Psi_\Omega$}}
    \put(39.5, 26.25){\color{graydiagram}{$\Gamma$}}
    \put(76, 3){\footnotesize $\mathcal{G}^T$}
    \put(76, 22.75){\footnotesize $\tilde{\mathcal{F}}^T$}
    \put(87.25, 19.5){\footnotesize $\mathcal{L}(\tilde{\mathcal{F}}^T, \mathcal{G}^T)$}
    \put(18, 52){\footnotesize $\mathcal{P}^Q$}
    \put(38.5, 67){\color{bluediagram}{$\Psi_\Omega$}}
    \put(38.5, 47.5){\color{reddiagram}{$\Phi_\Theta$}}
    \put(76, 62.5){\footnotesize $\mathcal{G}^Q$}
    \put(76, 43){\footnotesize $\mathcal{F}^Q$}
    \put(87.25, 59){\footnotesize $\mathcal{L}(\mathcal{F}^Q, \mathcal{G}^Q)$}

    \end{overpic}

    \vspace{-1.5mm}
    \caption{
    Overview of~\acronym.
    Top: A Query point cloud is being fed to both teacher and student network. We obtain distinctive 3D local descriptor from (frozen) teacher which guide the student descriptors via proposed distillation loss.
    Bottom: Rather then computing the teacher features for Target objects, we leverage the learning via correspondences. We transfer the features from query to target point cloud using the ground-truth transformation available at training time. By leveraging this, we significantly reduce the memory and compute overhead.
    }
    \label{fig:diagram}
\end{figure}
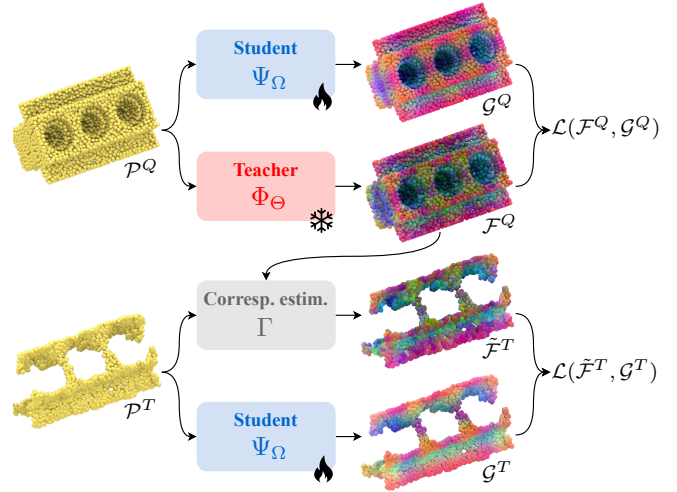

%% file: main/sections/5_results.tex
\section{Experiments}\label{sec:experiments}

\subsection{Experimental setup}

We validate \acronym with three experiments:
(i) a direct comparison against GeDi~\cite{poiesi2023gedi},
(ii) a detailed analysis of \acronym's key components, and
(iii) a comparison with state-of-the-art zero-shot pose estimators by integrating \acronym into FreeZe~\cite{caraffa2024freeze}.
During preprocessing, we randomly sample 4k points from $\mathcal{P}^Q$ and $\mathcal{P}^T$, and apply statistical outlier removal~\cite{open3d} to $\mathcal{P}^T$ to reduce depth noise.
Both point clouds are normalized by $\mathcal{P}^Q$'s diameter to promote scale invariance.
In (i) and (ii), we perform feature extraction using only the geometric encoder, while in (iii) we integrate \acronym with FreeZe's vision encoder.
In (i) and (iii), we use both ground-truth and SAM-6D~\cite{lin2023sam6d} segmentation masks to ensure fairness and compliance with the zero-shot assumption.
In (ii), we perform object localization using ground-truth segmentation masks to focus the analysis on feature quality without localization errors.
GeDi extracts 32-dimensional point-level features from local neighborhoods covering 30\% of $Q$'s diameter.
After feature extraction, we perform feature matching using the Open3D's RANSAC and ICP implementations.
We run 100k iterations for RANSAC and 1k iterations for ICP.
The student model is trained using eight NVIDIA A100 GPUs, while evaluation is performed on a single NVIDIA A40 GPU.
Note that the peak memory usage of \acronym when processing a point cloud with 5k points is only 717MB, making it suitable for deployment on edge devices with limited resources.
The student architecture is adapted from PTV3~\cite{wu2024ptv3}, using the same model settings but modifying the decoder output channels from 64 to 32 to match the teacher feature dimensionality. The model is trained using the AdamW optimizer with a cosine learning rate scheduler.


\vspace{1mm}
\noindent\textit{Training dataset.}
We train \acronym on the MegaPose-ShapeNetCore~\cite{labbe2022megapose} dataset, which contains one million synthetic images generated by rendering various 3D objects against random background scenes.
The dataset provides ground-truth segmentation masks, which we use to localize target objects.


\vspace{1mm}
\noindent\textit{Evaluation datasets.}
We validate \acronym on the BOP Benchmark datasets with publicly available ground-truth 6D poses: LM-O~\cite{lm}, T-LESS~\cite{tless}, TUD-L~\cite{tudl}, IC-BIN~\cite{icbin}, and YCB-V~\cite{ycbv}.
This benchmark presents several challenges, comprising a diverse set of everyday and industrial objects, both textured and texture-less objects, symmetries, varying levels of occlusion, and different lighting conditions.

\vspace{1mm}
\noindent\textit{Evaluation metrics.}
We evaluate the accuracy of \acronym using the ratio-of-nearest points (RON)~\cite{Zhou2022} and feature-matching recall (FMR)~\cite{deng2018ppfnet} metrics.
RON measures the percentage of query points with a nearest neighbor in the target object within a given distance threshold $\tau_1$.
FMR measures the percentage of points with RON greater than $\tau_2$.
During validation, we set $\tau_1 = 3\%$ of $Q$'s diameters and $\tau_2 = 5\%$.
Note that both metrics do not require registration, providing a direct assessment of feature quality.
To evaluate the 6D pose estimation accuracy, we use average recall (AR)~\cite{2020_bop_report}, as in the BOP Benchmark.
We evaluate the efficiency of \acronym by measuring the average processing time per image in seconds.

\subsection{Quantitative results}\label{sec:quant_results}

\input{main/tables/teacher}
\noindent\textit{GeDi vs.~\acronym.}
Tab.~\ref{tab:teacher} compares GeDi~\cite{poiesi2023gedi} (teacher, Rows~1-12) and \acronym (student, Rows~13-24) on the BOP Benchmark using different types of segmentation masks.
When using ground-truth segmentation masks, \acronym significantly outperforms GeDi by +4.1 RON, +11.8 FMR, and +3.9 AR, while being 170 times faster (Row~25).
\acronym outperforms GeDi in RON and FMR scores across all datasets but shows a slight drop in AR on TUD-L\cite{tudl} and YCB-V\cite{ycbv}.
This difference arises because RON and FMR directly assess feature quality, whereas AR is computed post registration, where RANSAC's robustness to outliers can obscure the true contribution of the features.
On T-LESS~\cite{tless}, \acronym achieves a substantial +22.9 AR improvement over the teacher.
We attribute this significant gain to the specific type of 3D models: unlike other datasets, T-LESS provides CAD models of industrial objects with fine-grained geometric details and sharp edges.
As shown in Fig.~\ref{fig:features}, \acronym excels in this setting, producing features that are smoother that the teacher's and better capture the underlying geometry.
When using SAM-6D zero-shot segmentation masks, \acronym outperforms the teacher for all three metrics by +2.7 RON, +9.4 FMR and +1.7 AR (Row~26).
This setting is more challenging than the former because zero-shot segmentation masks may have inaccuracies along object contours, can miss some parts of the object, and can include parts of other objects or background, while during distillation we use ground-truth segmentation masks.
Despite this, \acronym proves to be robust to the domain gap, achieving a superior feature quality compared to the teacher (better RON and FMR) and a superior AR.
Beyond improved accuracy, \acronym also achieves superior computational efficiency, extracting features at more than 50 frames per second. 
Given that each image contains an average of five target objects, this implies that \acronym can process approximately 260 point clouds per second.
Processing times (rightmost column) depends on the number of object instances per image, ranging from 3ms for a single instance in TUD-L to 34ms for many instances in IC-BIN~\cite{icbin}.
In summary, \acronym combines high accuracy and low latency, making it suitable for robotics applications.

\input{main/figures/features/features}

\vspace{1mm}
\noindent\textit{Ablation study.}
Tab.~\ref{tab:ablation} analyses the impact of three key components in \acronym: supervision type (Col.~1), loss function design (Cols.~2-5), and data augmentation strategy (Col.~6-8).
Row~1 is our baseline, which uses supervision only from query objects, Eq.~\ref{eq:v2} as loss, and no data augmentation.
Rows~2-8 explore alternative designs, while Row~9 is \acronym's final configuration.
We evaluate each variant on the BOP Benchmark and report its improvement over the baseline in the rightmost column.
We draw three main insights for these experiments.
First, the learning via correspondence paradigm introduced in Sec.~\ref{sec:distill_corresp} improves performance considerably (+4.7~AR, Row~5) by introducing supervision from partial and noisy target objects with the same memory requirements.
Second, the proposed loss (Eq.~\ref{eq:v2}) outperforms both traditional designs (+3.4~AR against MSE in Row~2, +3.1~AR against Huber in Row~3) and Eq.~\ref{eq:v1} (+8.5~AR, Row~4).
This indicates that effective distillation of frozen GeDi features requires a careful loss design, and that reducing the contribution of points with a large regression error is beneficial.
Third, data augmentation promote generalization (Row~9 achieves +1.3~AR over Row~5).
These ablation experiments confirm that integrating correspondence-based supervision, a tailored loss, and data augmentation consistently enhances performance.

\input{main/tables/ablation}

\input{main/tables/freeze}

\vspace{1mm}
\noindent\textit{\acronym and FreeZe integration.}
Tab.~\ref{tab:freeze} assesses the 6D pose estimation performance obtained when integrating \acronym in FreeZe.
We compare against state-of-the-art zero-shot 6D pose estimators on the same datasets as Tab.~\ref{tab:teacher}, evaluating AR and run-time.
We compare the performance of \acronym with SAM-6D~\cite{lin2023sam6d}, FoundationPose~\cite{wen2023foundationpose}, and the original version of FreeZe, which are the state-of-the-art zero-shot methods.
Rows~1-3 present publicly available results from the BOP Benchmark leaderboard~\cite{bopleaderboard}.
Rows~4-5 present the reproduced results of FreeZe, both with and without symmetry-aware pose refinement (SAR).
Rows~6-7 show performance when replacing FreeZe's geometric encoder GeDi~\cite{poiesi2023gedi} with \acronym.
To ensure a fair comparison, all methods use the SAM-6D zero-shot segmentor for object localization except for Rows~8-9 where ground-truth segmentation masks are used.
For experiments using ground-truth segmentation masks, \acronym achieves the same AR of FreeZe, while reducing the inference time by 2.7 seconds (Row~9).
When using SAM-6D segmentation masks, \acronym reaches 70.4 AR compared to FreeZe's 71.5, yet it runs 5.2 seconds faster.
In contrast, the slight performance drop is mainly due to noise in the predicted segmentation masks. 
Since \acronym is trained with ground-truth, noise-free segmentation masks, the introduction of noisy SAM-6D segmentation masks creates a domain gap that leads to suboptimal pose estimation, although \acronym still maintains better feature quality and faster processing.


\subsection{Qualitative results}

\input{main/figures/qualitatives}

Figure~\ref{fig:qualitative} presents qualitative results with a clear layout: the first column shows the input images, the second displays predictions from \acronym, and the third contains those from GeDi. 
For clarity, the predicted poses are overlaid on the input images and the RON is indicated for each case. 
Each row corresponds to an example drawn from a different dataset, addressing various challenges such as occlusions (a, d, e), object symmetries (b), partial views (c), and the presence of multiple instances of the same object type (d).
In examples (a) and (e), the toy cat is partially occluded by a watering can (from LM-O) and clipper tool (from YCBV) is obscured by a food can. 
GeDi fails in these cases due to the insufficient visible structure needed to reliably form accurate LRFs. This leads to incorrect pose predictions. 
In contrast, \acronym excels by directly distilling robust local geometric descriptors from the available cues. 
This targeted learning helps \acronym to accurately recover the pose even when only limited geometric information is present.
In (d), a juice box is heavily occluded by cups in a challenging bin picking scenario involving multiple instances. 
Although the teacher’s prediction (with a RON of 0.01) appears quantitatively favorable, its reliance on sparse LRF cues leads to unreliable pose estimates under these adverse occlusions. 
In this case, \acronym achieves a RON of 0.02 by extracting and utilizing richer local geometric details, thereby successfully recovering the correct pose despite the limited and cluttered geometric information.
In (b), a symmetrical square electrical component is the target. \acronym learns fine discriminative features even in the presence of symmetry, overcoming the ambiguities that affect the GeDi's LRF. By focusing on fine patterns and slight asymmetries within a geometrically symmetrical structure, \acronym achieves an accurate pose prediction.
In summary, for all cases, \acronym generates more distinctive descriptors compared to GeDi (higher RON) and successfully recovers the correct 6D poses.


%% file: main/tables/teacher.tex
\begin{table}[t]
\vspace{2.5mm}
\renewcommand{\arraystretch}{0.9}
\centering
\tabcolsep 5pt
\caption{
Comparison of GeDi and \acronym on the BOP Benchmark in terms of feature quality (RON, FMR) and pose estimation accuracy (AR) with ground-truth (GT) and zero-shot (ZS) segmentation masks.
Average feature extraction time per image is reported in seconds.
Rows~25-26 highlight \acronym improvement over GeDi.
}
\label{tab:teacher}

\vspace{-2mm}
\resizebox{\columnwidth}{!}{%
    \begin{tabular}{rlcc|rrrr}

    \toprule
    & Dataset & GT & ZS & RON {\color{gray}$\uparrow$} & FMR {\color{gray}$\uparrow$} & AR {\color{gray}$\uparrow$} & Run-time {\color{gray}$\downarrow$} \\
    \midrule
    \multicolumn{8}{c}{GeDi~\cite{poiesi2023gedi}} \\
    {\color{gray}\scriptsize 1} &\multirow{2}{*}{LM-O} & \checkmark & & 1.8 & 8.8 & 66.9 & \multirow{2}{*}{3.198} \\
    {\color{gray}\scriptsize 2} & & & \checkmark & 3.5 & 7.4 & 60.7 & \\
    \cmidrule{2-8}
    {\color{gray}\scriptsize 3} & \multirow{2}{*}{T-LESS} & \checkmark & & 0.4 & 1.0  & 41.6 & \multirow{2}{*}{3.600} \\
    {\color{gray}\scriptsize 4} & & & \checkmark & 0.4 & 0.7  & 33.0 & \\
    \cmidrule{2-8}
    {\color{gray}\scriptsize 5}& \multirow{2}{*}{TUD-L}  & \checkmark &    & 8.2 & 75.5 & 96.8 & \multirow{2}{*}{0.604} \\
    {\color{gray}\scriptsize 6} & & & \checkmark & 6.6 & 58.6 & 89.5 & \\
    \cmidrule{2-8}
    {\color{gray}\scriptsize 7}&\multirow{2}{*}{IC-BIN} & \checkmark &    & 1.0 & 11.6 & 49.9 & \multirow{2}{*}{6.126} \\
    {\color{gray}\scriptsize 8}  &                  & & \checkmark & 0.6 & 1.8  & 39.5 & \\
    \cmidrule{2-8}
    {\color{gray}\scriptsize 9}&\multirow{2}{*}{YCB-V}  & \checkmark &    & 3.6 & 25.8 & 59.1 & \multirow{2}{*}{2.892} \\
    {\color{gray}\scriptsize 10}&                    & & \checkmark & 3.2 & 23.7 & 53.0 & \\
    \cmidrule{2-8}
    {\color{gray}\scriptsize 11}& \multirow{2}{*}{Avg}    & \checkmark &    & 3.0 & 24.5 & 62.9 & \multirow{2}{*}{3.284} \\
    {\color{gray}\scriptsize 12}&                     & & \checkmark & 2.9 & 18.5 & 55.1 & \\
    \midrule
    \multicolumn{8}{c}{\acronym (ours)} \\
    {\color{gray}\scriptsize 13}& \multirow{2}{*}{LM-O}   & \checkmark &    & 4.3 & 27.2 & 67.2 & \multirow{2}{*}{0.021} \\
    {\color{gray}\scriptsize 14}&                      & & \checkmark & 1.7 & 23.5 & 60.4 & \\
    \cmidrule{2-8}
    {\color{gray}\scriptsize 15}& \multirow{2}{*}{T-LESS} & \checkmark &    & 2.1 & 15.6 & 64.5 & \multirow{2}{*}{0.021} \\
    {\color{gray}\scriptsize 16}&                    & & \checkmark & 1.6 & 11.6 & 44.7 & \\
    \cmidrule{2-8}
    {\color{gray}\scriptsize 17}& \multirow{2}{*}{TUD-L}  & \checkmark &    & 16.3 & 77.8 & 93.2 & \multirow{2}{*}{0.003} \\
    {\color{gray}\scriptsize 18}&                     & & \checkmark & 12.8 & 67.3 & 83.0 & \\
    \cmidrule{2-8}
    {\color{gray}\scriptsize 19}& \multirow{2}{*}{IC-BIN} & \checkmark &    & 3.7 & 18.5 & 50.5 & \multirow{2}{*}{0.034} \\
    {\color{gray}\scriptsize 20}&                    & & \checkmark & 4.5 & 2.6  & 43.4 & \\
    \cmidrule{2-8}
    {\color{gray}\scriptsize 21}&  \multirow{2}{*}{YCB-V}  & \checkmark &    & 9.1 & 42.3 & 58.8 & \multirow{2}{*}{0.015} \\
    {\color{gray}\scriptsize 22}&                     & & \checkmark & 7.4 & 34.8 & 52.3 & \\
    \cmidrule{2-8}
    {\color{gray}\scriptsize 23} & \multirow{2}{*}{Avg} & \checkmark & & \textbf{7.1} & \textbf{36.3} & \textbf{66.8} & \multirow{2}{*}{0.019} \\
    {\color{gray}\scriptsize 24}&                     & & \checkmark & \textbf{5.6} & \textbf{27.9} & \textbf{56.8} & \\
    \midrule
    {\color{gray}\scriptsize 25} & \multirow{2}{*}{$\Delta$} & \checkmark &  & \color{forestgreen}\textbf{+4.1} & \color{forestgreen}\textbf{+11.8} & \color{forestgreen}\textbf{+3.9} & \multirow{2}{*}{\color{forestgreen}\textbf{173$\times$}}\\ 
    {\color{gray}\scriptsize 26} & & & \checkmark & \color{forestgreen}\textbf{+2.7} & \color{forestgreen}\textbf{+9.4} & \color{forestgreen}\textbf{+1.7} & \\
    \bottomrule  
    \end{tabular}
    }
    \vspace{-1.5mm}
\end{table}

%% file: main/figures/features/features.tex
\begin{figure}[t!]
    \vspace{-1.5mm}
    \centering
    \begin{overpic}[trim=100 0 100 0, width=0.31\columnwidth]{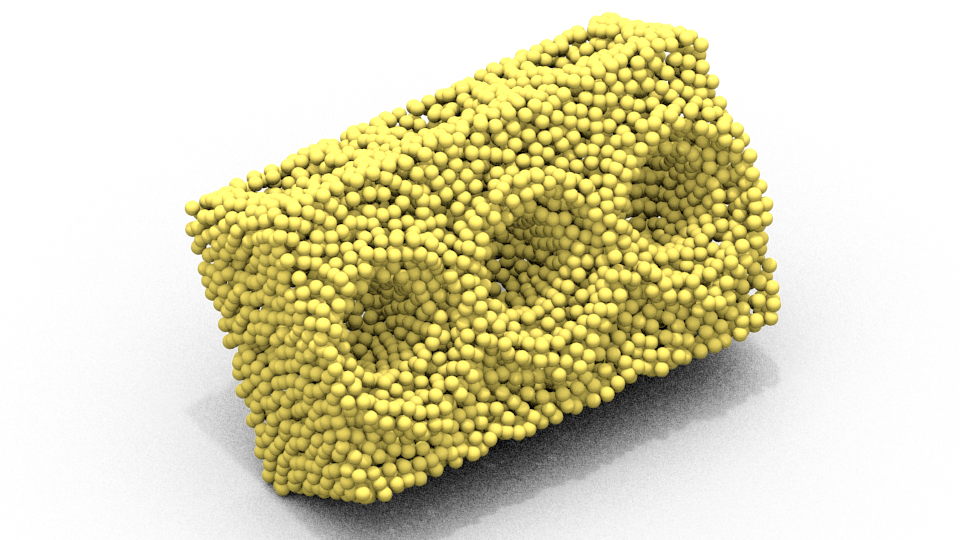}
        \put(80,5){$\mathcal{P}^Q$}
    \end{overpic}
    \begin{overpic}[trim=100 0 100 0, width=0.31\columnwidth]{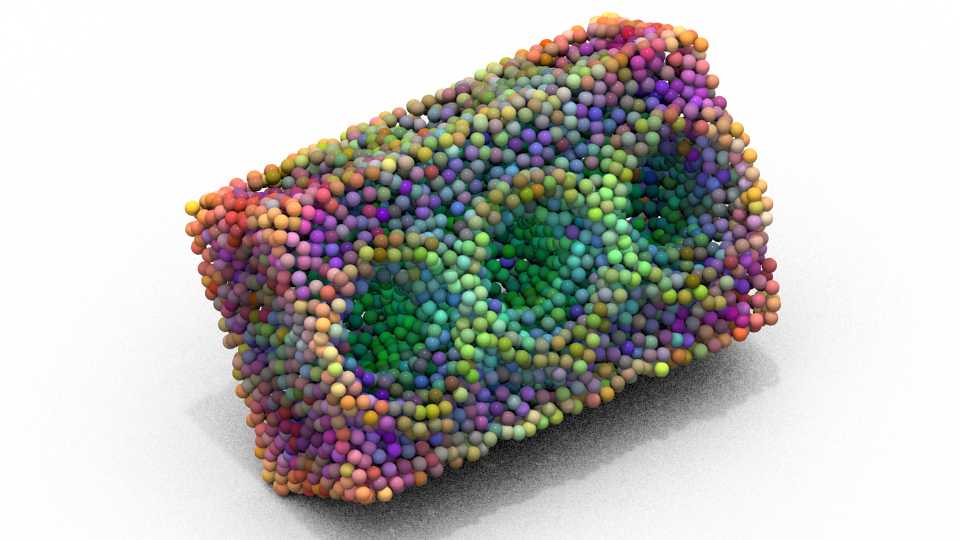}
        \put(80,5){$\mathcal{F}^Q$}
    \end{overpic}
    \begin{overpic}[trim=100 0 100 0, width=0.31\columnwidth]{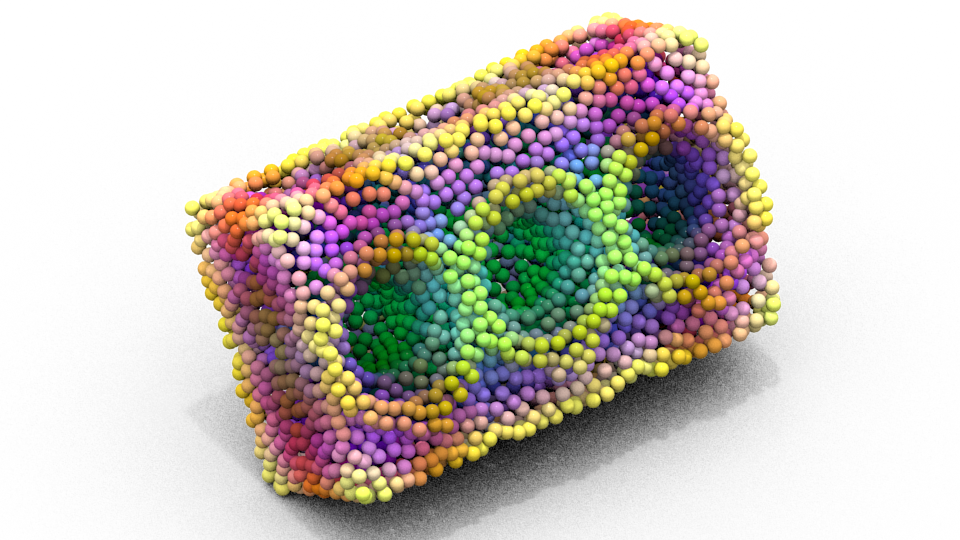}
        \put(80,5){$\mathcal{G}^Q$}
    \end{overpic}


    \vspace{-2.5mm}
    \caption{
    Comparison between teacher (center) and student (right) features on a query object from T-LESS (left). Colors represent PCA-reduced features. Student features $\mathcal{G}^Q$ are smoother and less noisy than teacher features $\mathcal{F}^Q$.
    }
    \label{fig:features}
    \vspace{-5mm}
\end{figure}

%% file: main/tables/ablation.tex
\begin{table*}[t]
    \vspace{0.8mm}
    \centering
    \tabcolsep 4pt
    \caption{
    Ablation analysis on three key components: supervision strategy, loss function, and data augmentation.
    The rightmost column indicates each variant's improvement over the baseline (Row 1).
    Row~9 shows the final configuration of \acronym.
    }
    \label{tab:ablation}

    \vspace{-2mm}
    \resizebox{\textwidth}{!}{%
    \begin{tabular}{rc|cccc|ccc|ccccccc}
        \toprule
        & Target obj. & \multicolumn{4}{c|}{Loss function} & \multicolumn{3}{c|}{Data augmentation} & \multicolumn{6}{c}{BOP Benchmark} \\
        & supervision & MSE & Huber & Eq.~\ref{eq:v1} & Eq.~\ref{eq:v2} & Rotation & Jitter & Point drop. & 
        LM-O & T-LESS & TUD-L & IC-BIN & YCB-V & Avg & $\Delta$ \\
        \midrule
        {\color{gray} \scriptsize 1} & & & & & \checkmark & & & & 64.5 & 53.7 & 88.5 & 46.1 & 51.3 & 60.8 & - \\
        \midrule
        {\color{gray} \scriptsize 2} & \checkmark & \checkmark & & & & & & & 59.5 & 60.2 & 86.4 & 47.6 & 56.7 & 62.1 & \color{forestgreen}{+1.3} \\
        {\color{gray} \scriptsize 3} & \checkmark & & \checkmark & & & & & & 61.3 & 59.9 & 85.5 & 48.3 & 56.8 & 62.4 & \color{forestgreen}{+1.6} \\
        {\color{gray} \scriptsize 4} & \checkmark & & & \checkmark & & & & & 51.2 & 54.5 & 77.2 & 45.8 & 56.1 & 57.0 & \color{red}{-3.8} \\
        {\color{gray} \scriptsize 5} & \checkmark & & & & \checkmark & & & & 67.5 & 62.2 & 94.1 & 46.1 & 57.8 & 65.5 & \color{forestgreen}{+4.7}\\
        {\color{gray} \scriptsize 6} & \checkmark & & & & \checkmark & \checkmark & & & 68.9 & 63.1 & 94.2 & 47.1 & 58.0 & 66.3 & \color{forestgreen}{+5.5} \\
        {\color{gray} \scriptsize 7} & \checkmark & & & & \checkmark &  & \checkmark  & & 67.8 & 63.4 & 93.6 & 48.5 & 58.2 & 66.3 & \color{forestgreen}{+5.5} \\
        {\color{gray} \scriptsize 8} & \checkmark & &  &  & \checkmark &  &   & \checkmark & 67.7 & 64.0 & 93.5 & 48.7 & 58.5 & 66.5 & \color{forestgreen}{+5.7} \\
        \midrule
        {\color{gray} \scriptsize 9} & \checkmark & & & & \checkmark & \checkmark & \checkmark & \checkmark & 67.2 & 64.5 & 93.2 & 50.5 & 58.8 & \textbf{66.8}  & \color{forestgreen}{\textbf{+6.0}} \\
        \bottomrule
    \end{tabular}
    }
\end{table*}

%% file: main/tables/freeze.tex
\begin{table*}[t]
\centering
\caption{
Results on the BOP Benchmark in terms of AR and run-time with ground-truth (GT) and zero-shot (ZS) segmentation masks.
}
\label{tab:freeze}

\vspace{-2mm}
\resizebox{\textwidth}{!}{%
\begin{tabular}{rlcc|cccccc|rr}
\toprule

    & \multirow{2}{*}{Method} & \multicolumn{2}{c|}{Segm.} & \multicolumn{6}{c|}{AR {\color{gray} $\uparrow$}} & \multicolumn{2}{c}{Run-time (s) {\color{gray} $\downarrow$}} \\

    & & GT & ZS & LM-O & T-LESS & TUD-L & IC-BIN & YCB-V & Avg & Avg & GPU \\
    \toprule
    %
    %
    {\color{gray} \scriptsize 1} & SAM-6D~\cite{lin2023sam6d} & & \checkmark & 69.9 & 51.5 & 90.4 & 58.8 & 84.5 & 71.0 & 4.5 & RTX3090 \\
    %
    %
    {\color{gray} \scriptsize 2} & FoundationPose~\cite{wen2023foundationpose} & & \checkmark & 75.6 & 64.6 & 92.3 & 50.8 & 88.9 & 74.4 & 35.9 & A100 \\
    %
    %
    {\color{gray} \scriptsize 3} & FreeZe~\cite{caraffa2024freeze} & & \checkmark & 71.6 & 53.1 & 94.9 & 54.5 & 84.0 & 71.6 & 12.7 & A40 \\
    \midrule
    {\color{gray} \scriptsize 4} & FreeZe~\cite{caraffa2024freeze} w/o SAR & & \checkmark & 71.4 & 53.0 & 94.8 & 51.5 & 83.5 & 70.9 & 10.8 & A40 \\
    {\color{gray} \scriptsize 5} & FreeZe~\cite{caraffa2024freeze} & & \checkmark & 71.4 & 53.0 & 94.8 & 54.3 & 83.9 & 71.5 & 13.7 & A40 \\
    \rowcolor{tableazure} {\color{gray} \scriptsize 6} & FreeZe w/ \acronym  w/o SAR (ours) & & \checkmark &69.6 & 53.5 & 93.1 & 49.1 & 79.7 & 69.0 & 6.4 & A40 \\
    \rowcolor{tableazure} {\color{gray} \scriptsize 7} & FreeZe w/ \acronym (ours) & & \checkmark & 69.6 & 53.5 & 93.1 & 52.3 & 83.4 & 70.4 & 8.5 & A40 \\
    \midrule
    {\color{gray} \scriptsize 8} & FreeZe~\cite{caraffa2024freeze} w/o SAR & \checkmark & & 80.2 & 75.1 & 98.6 & 53.8 & 90.2 & 79.6 & 8.7 & A40 \\
    \rowcolor{tableazure} {\color{gray} \scriptsize 9} & FreeZe w/ \acronym w/o SAR (ours) & \checkmark & & 80.5 & 75.0 & 98.7 & 54.2 & 89.4 & 79.6 & 6.0 & A40 \\
    \bottomrule
\end{tabular}
}
\end{table*}

%% file: main/figures/qualitatives.tex
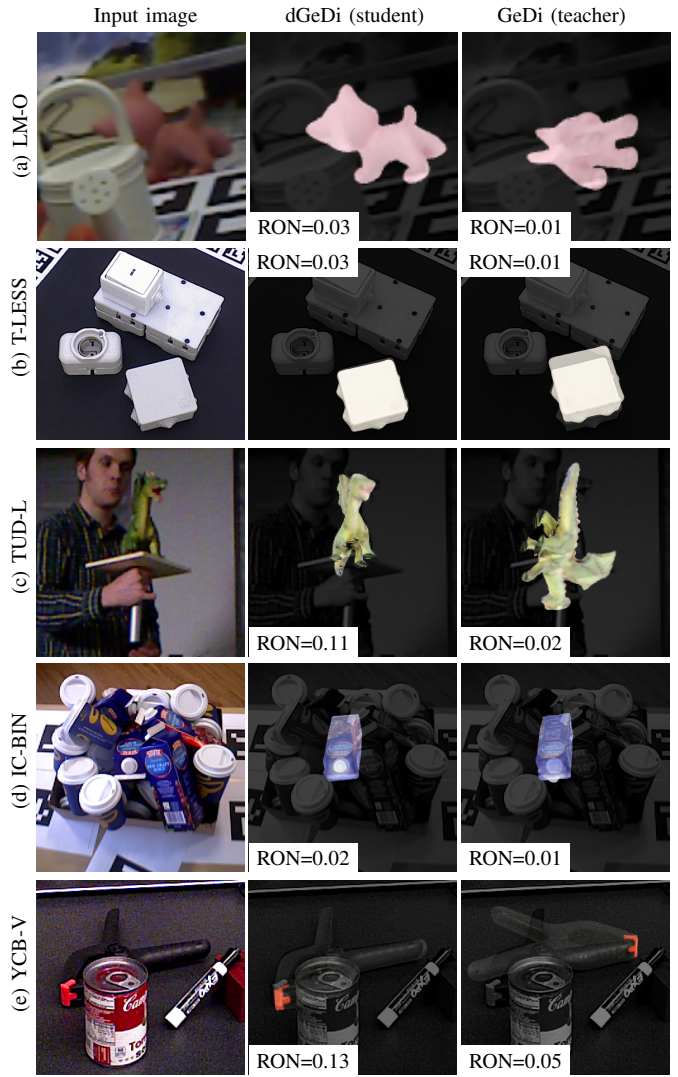
\begin{figure}[t]
\vspace{5.22mm}
\centering
\begin{tabular}{@{}c@{\,}c@{\,}c@{}}

    \begin{overpic}[width=0.32\columnwidth]{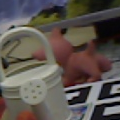}
        \put(-12,30){\rotatebox{90}{\footnotesize (a) LM-O}}
        \put(27,105){\footnotesize Input image}
    \end{overpic} &
    \begin{overpic}[width=0.32\columnwidth]{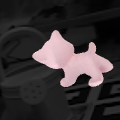}
        \put(17,105){\rotatebox{0}{\footnotesize \acronym (student)}}
        \put(0, 3){\colorbox{white}{\footnotesize RON=0.03}}
    \end{overpic} &
    \begin{overpic}[width=0.32\columnwidth]{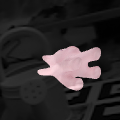}
        \put(17,105){\rotatebox{0}{\footnotesize GeDi (teacher)}}
        \put(0, 3){\colorbox{white}{\footnotesize RON=0.01}}
    \end{overpic} \\

    \begin{overpic}[width=0.32\columnwidth]{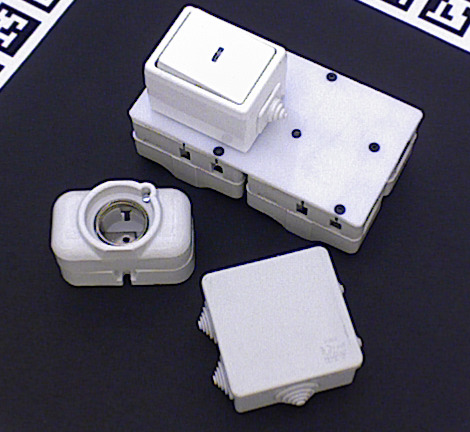}
        \put(-12,30){\rotatebox{90}{\footnotesize (b) T-LESS}}
    \end{overpic} &
    \begin{overpic}[width=0.32\columnwidth]{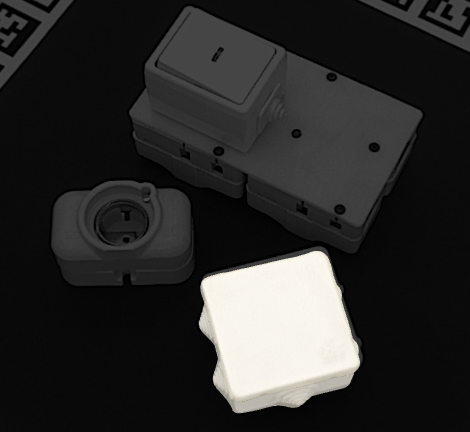}
    \put(0, 82){\colorbox{white}{\footnotesize RON=0.03}}
    \end{overpic} &
    \begin{overpic}[width=0.32\columnwidth]{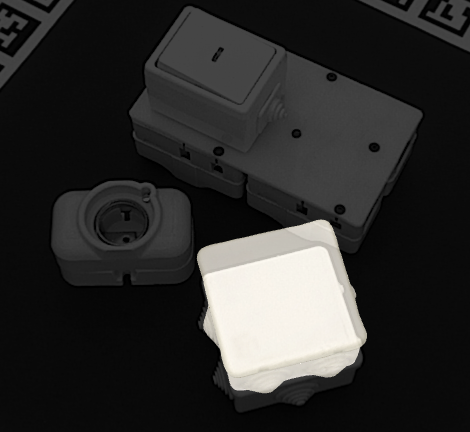}
    \put(0, 82){\colorbox{white}{\footnotesize RON=0.01}}
    \end{overpic} \\

    \begin{overpic}[width=0.32\columnwidth]{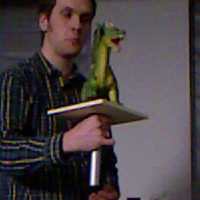}
        \put(-12,30){\rotatebox{90}{\footnotesize (c) TUD-L}}
    \end{overpic} &
    \begin{overpic}[width=0.32\columnwidth]{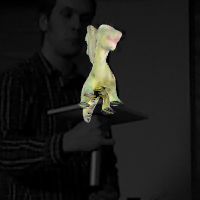}
    \put(0, 3){\colorbox{white}{\footnotesize RON=0.11}}
    \end{overpic} &
    \begin{overpic}[width=0.32\columnwidth]{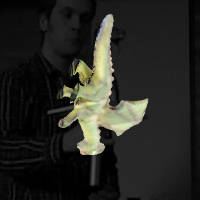}
    \put(0, 3){\colorbox{white}{\footnotesize RON=0.02}}
    \end{overpic} \\

    \begin{overpic}[width=0.32\columnwidth]{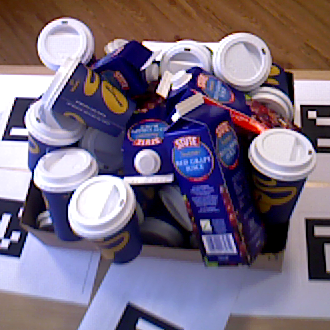}
        \put(-12,30){\rotatebox{90}{\footnotesize (d) IC-BIN}}
    \end{overpic} &
    \begin{overpic}[width=0.32\columnwidth] {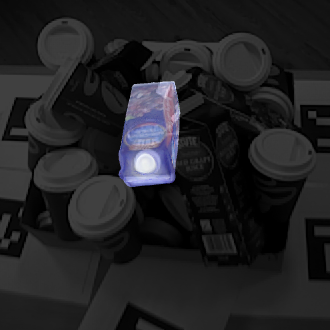}
    \put(0, 3){\colorbox{white}{\footnotesize RON=0.02}}
    \end{overpic} &
    \begin{overpic}[width=0.32\columnwidth] {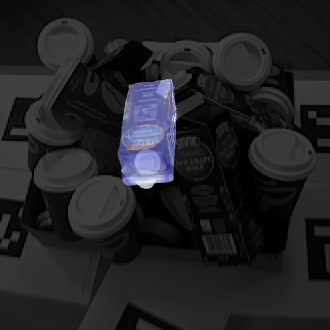}
    \put(0, 3){\colorbox{white}{\footnotesize RON=0.01}}
    \end{overpic} \\

    \begin{overpic}[width=0.32\columnwidth]{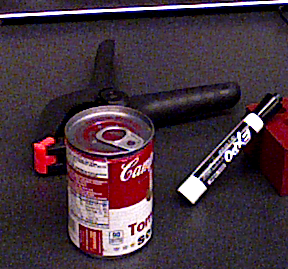}
        \put(-12,30){\rotatebox{90}{\footnotesize (e) YCB-V}}
    \end{overpic} &
    \begin{overpic}[width=0.32\columnwidth]{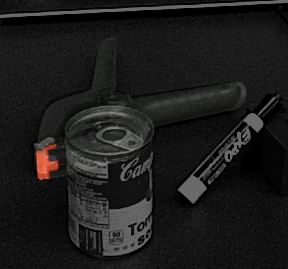}
    \put(0, 3){\colorbox{white}{\footnotesize RON=0.13}}
    \end{overpic} &
    \begin{overpic}[width=0.32\columnwidth]{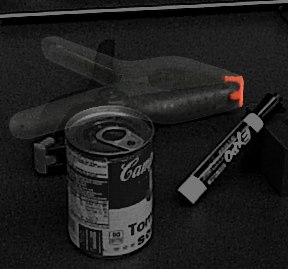}
    \put(0, 3){\colorbox{white}{\footnotesize RON=0.05}}
    \end{overpic} \\

\end{tabular}

\vspace{-0.5mm}
\caption{
Qualitative results on the BOP Benchmark (left) comparing \acronym (center) with GeDi (right).
We overlay the object's 3D model transformed according to the predicted pose on the greyscale input image for better contrast.
Key challenges: occlusions (a, d, e), object symmetry (b), partial view (c), multiple instances (d).
}
\label{fig:qualitative}
\vspace{-5.5mm}
\end{figure}

%% file: main/sections/6_conclusion.tex
\section{Conclusions}\label{sec:conclusion}

In this work, we explored knowledge distillation for 3D local descriptors, aiming to retain the effectiveness of GeDi while significantly improving its efficiency. 
By training an efficient student model to regress GeDi descriptors, we addressed key challenges such as the need for large-scale point cloud datasets, the transfer of rotation and scale invariance, and handling low inlier ratios during distillation. 
Our approach explicitly enforces consistency between query and target descriptors, enhancing robustness to occlusions and partial observations. 
Experimental results demonstrate that our distilled model achieves a substantial reduction in inference time while maintaining competitive performance with alternative methods. 
This brings zero-shot 6D pose estimation closer to real-time feasibility, making it more viable for practical applications. 
To the best of our knowledge, this is the first attempt to apply knowledge distillation to 3D local descriptors, opening new possibilities for efficient geometric reasoning in real-world scenarios.

\noindent\emph{Limitations}.
Our work focuses solely on improving the efficiency of geometric feature extraction within the FreeZe pipeline.
While this enhances performance, other components, such as registration, remain computational bottlenecks.

\noindent\emph{Future works}.
We plan to explore strategies to accelerate these remaining modules, further optimizing the entire 6D pose estimation pipeline for real-time applications.